\documentclass[letterpaper, 10 pt, conference]{ieeeconf}  % Comment this line out if you need a4paper

\usepackage{array}
\usepackage{xcolor}
\usepackage{amsmath}

\usepackage{amsthm}
\usepackage{amssymb}
\usepackage[font={small}]{caption}
\usepackage{subcaption}
\usepackage{graphicx}
\usepackage{url}
\usepackage[noadjust]{cite}
\newtheorem{remark}{Remark}
 
\IEEEoverridecommandlockouts                              % This command is only needed if 
\usepackage{bbm}
\usepackage{dsfont}
\usepackage{hyperref}

\title{\LARGE \bf
Preferenced Oracle Guided Multi-mode Policies \\
for Dynamic Bipedal Loco-Manipulation
}

\author{
    Prashanth Ravichandar*, Lokesh Krishna*, Nikhil Sobanbabu, and Quan Nguyen % <-this % stops a space
    \thanks{* The authors contributed equally.}
    \thanks{All authors are associated with the Dynamic Robotics and Control  Laboratory, University of Southern California, Los Angeles, CA 90089, USA 
    {\tt\small pr39760@usc.edu, lkrajan@usc.edu,  ns\_562@usc.edu, quann@usc.edu}}
}

\begin{document}

\maketitle
\thispagestyle{empty}
\pagestyle{empty}

%%%%%%%%%%%%%%%%%%%%%%%%%%%%%%%%%%%%%%%%%%%%%%%%%%%%%%%%%%%%%%%%%%%%%%%%%%%%%%%%
\begin{abstract}

Dynamic loco-manipulation calls for effective whole-body control and contact-rich interactions with the object and the environment. Existing learning-based control synthesis relies on training low-level skill policies and explicitly switching with a high-level policy or a hand-designed finite state machine, leading to quasi-static behaviors. In contrast, dynamic tasks such as soccer require the robot to run towards the ball, decelerate to an optimal approach to dribble, and eventually kick a goal—a continuum of smooth motion. To this end, we propose Preferenced Oracle Guided Multi-mode Policies (OGMP) to learn a single policy mastering all the required modes and preferred sequence of transitions to solve uni-object loco-manipulation tasks. We design hybrid automatons as oracles to generate references with continuous dynamics and discrete mode jumps to perform a guided policy optimization through bounded exploration. To enforce learning a desired sequence of mode transitions, we present a task-agnostic preference reward that enhances performance. The proposed approach demonstrates successful loco-manipulation for tasks like soccer and moving boxes omnidirectionally through whole-body control. In soccer, a single policy learns to optimally reach the ball, transition to contact-rich dribbling, and execute successful goal kicks and ball stops. Leveraging the oracle's abstraction, we solve each loco-manipulation task on robots with varying morphologies, including HECTOR V1, Berkeley Humanoid, Unitree G1, and H1, using the same reward definition and weights. 

\end{abstract}

% We design oracles  as a closed-loop state-reference generator, viewing it as a hybrid automaton with continuous reference-generating dynamics and discrete mode jumps. 

%%%%%%%%%%%%%%%%%%%%%%%%%%%%%%%%%%%%%%%%%%%%%%%%%%%%%%%%%%%%%%%%%%%%%%%%%%%%%%%%
\section{Introduction}
\label{sec:intro}

Robot control for loco-manipulation requires tight integration of locomotion and manipulation to interact extensively with the object and environment. Traditionally, agile locomotion and dexterous manipulation have been studied individually, with the overlapping focus being robustness and recovery behaviors. In locomotion, the robot's contact with its environment is limited to the end-effector while maintaining balance and tracking a desired CoM reference over challenging terrains. In contrast, manipulation requires multiplicity in contact interactions with objects of distinct geometries and dynamics. For effective loco-manipulation, a control policy is thus required to stabilize the underactuated states of the robot and the object, leveraging its whole body to make and break contact to solve the task effectively. For loco-manipulation, model-based optimal control relies on a pre-planned contact schedule and knowledge of an object's physical properties \cite{li2023multi, pickingup}. Selecting an optimal sequence of contacts is combinatorial, typically addressed through sampling-based \cite{zhu2023efficient} or heuristic-based \cite{jorgensen202finding} planners, followed by trajectory optimization \cite{sleiman2023versatile}, with recent efforts toward combined optimization \cite{marcucci2024shortes}. Nonetheless, its online computational tractability remains challenging when augmenting object dynamics, highlighting an open problem in loco-manipulation control.

 \begin{figure}
	\centering
	\includegraphics[width=0.48\textwidth]{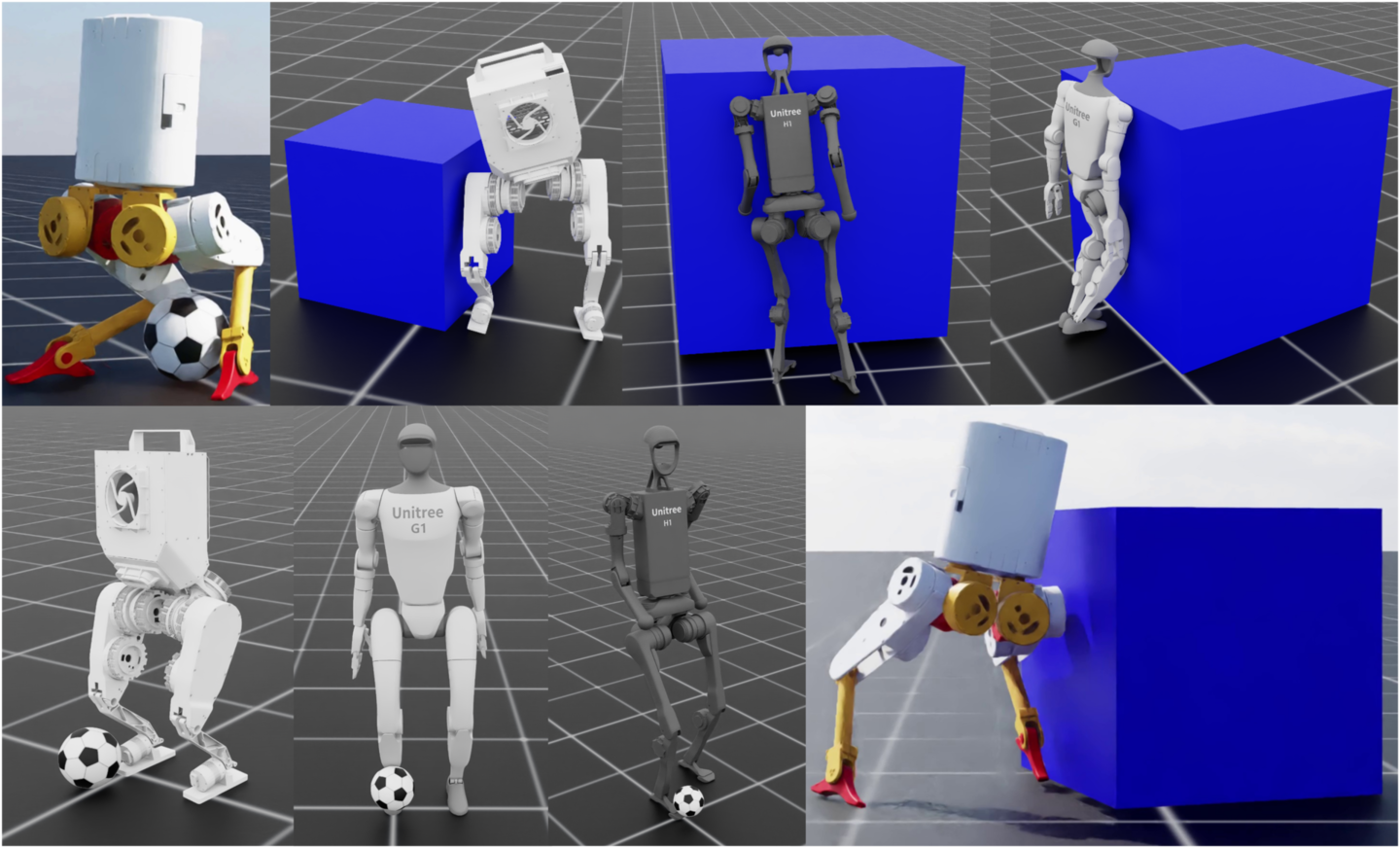}
	\caption{Dynamic contact-rich whole body loco-manipulation tasks on different bipedal and humanoid robots with Preferenced OGMP. Accompanying project website: \href{https://indweller.github.io/ogmplm/}{\color{blue}https://indweller.github.io/ogmplm/}}
	\label{fig:intro_fig}
 \vspace{-6mm}
\end{figure}

As a compelling alternative, Deep Reinforcement Learning (RL) for robot control scales to high-dimensional simulation models \cite{lee2020learning, cheng2024extreme, radosavovic2024real, tan2018sim, hwangbo2019learning, margolis2024rapid}. Optimizing a policy over stochastic dynamics allows introducing randomization, perturbation, and model uncertainty as seen in a real robot, thereby leading to robust controllers for legged locomotion in the wild \cite{siekmann2021blind,miki2022learning} and dexterous manipulation of objects with complex geometry \cite{chen2023visual}. For uni-object quadrupedal loco-manipulation, prior works leverage hierarchical frameworks with a versatile low-level locomotion policy and a high-level policy for arm-control \cite{dadiotis2025dynamicobjectgoalpushing} or navigation commands \cite{jeon2024learning}. Analogously, \cite{huang2022creating} proposes a high-level switching policy and multiple low-level skill policies for agile goalkeeping. With two independently trained policies and a hand-designed finite state machine, \cite{ji2023dribblebot} showcases dynamic quadrupedal dribbling and recovery. Similarly, \cite{dao2023simtoreallearninghumanoidbox} learns multiple low-level policies with a subsequent state-initialization strategy for various modes of operation and executes a predetermined sequence of policy switches for humanoid box loco-manipulation. A shared motif of the above methods is to leverage hierarchical control with a learned high-level policy or a hand-designed finite state machine with pre-trained low-level policies. Despite the promising empirical results, it remains unclear whether the hierarchy and learning of multiple policies are a fundamental necessity or a practical convenience for loco-manipulation. Moreover, the amount of task-specific engineering, combined with the lack of abstraction, limits the ability of these approaches to be adopted across different tasks and robots. Alternatively, \cite{haarnoja2024learning} learns multiple low-level skill policies but finally distills them into a single policy for humanoid soccer. Thus, a single policy network can approximate all the necessary skills, yet a multi-policy training strategy is adopted to 1) ease the reward shaping for each skill and 2) alleviate the local optima arising from undesirable skill transitions. In this work, we learn a single policy that masters multiple skills (or control modes) via guided policy optimization, prioritizing desired mode transitions while maintaining an abstraction extendable to different robots and tasks.
% we focus on learning a single policy mastering multiple skills (or control modes) through guided policy optimization with a preference for desired mode transitions while ensuring a level of abstraction that can be extended to different robots and tasks.   

Recently \cite{sleiman2025guided} proposed a task-agnostic guided policy optimization framework for multi-contact quadrupedal loco-manipulation. 
% A single policy per task is guided by a pre-computed demonstration from an offline multi-modal planner \cite{sleiman2023versatile} with continuous trajectories and contact schedules with adaptive phase dynamics to accommodate unforeseen discrepancies between the demonstration and policy rollout. 
A task-specific policy is guided by a pre-computed demonstration from an offline multi-modal planner \cite{sleiman2023versatile}, featuring continuous trajectories and contact schedules with adaptive phase dynamics to handle discrepancies between demonstration and rollout.
While \cite{sleiman2025guided} is tractable for motion with sparse user-defined contact affordances \cite{sleiman2023versatile}, the scalability to contact-dense bipedal dribbling is not straightforward, highlighting the need for emergent behaviors through exploration. Concurrent work \cite{krishna2024ogmp} proposes Oracle Guided Multi-mode Policies (OGMP) as a framework for structured exploration by bounding the permissible states to a local neighborhood of a coarse state reference generated by a closed-loop oracle. In \cite{krishna2024ogmp}, a single policy is trained for solving agile bipedal parkour with control modes like leap for gaps, jump for blocks, and pace for flat terrain, traversing arbitrary parkour tracks. Unlike other approaches, the mode transitions in OGMP are implicit and emergent without relying on high-level modules, leading to effective transition maneuvers. While OGMP was validated with oracles using robot state feedback and exogenous time-varying inputs (like height-map), the design of oracles to accommodate external object dynamics is unexplored. Unlike parkour, where terrain geometry uniquely determines the mode transitions, soccer involves multiple possible mode sequences to solve the task, which is the focus of this work. We first introduce a hybrid automata perspective to design multi-mode oracles with continuous reference-generating dynamics and discrete mode switches. With such an oracle, we propose a Preferenced OGMP to synthesize multi-mode control favoring desirable mode transitions. The key contribution of our paper is two-fold:
\begin{enumerate}
    \item  We formalize oracle design using a hybrid automata perspective to model the multi-mode reference generation. By designing a single oracle, we solve different uni-object loco-manipulation tasks such as omnidirectional moving box and soccer variants on multiple bipeds and humanoid robots including HECTOR, Berkeley Humanoid, Unitree G1 and H1 (Fig. \ref{fig:intro_fig}). 
    
    \item We introduce a task-agnostic preference reward that directs the policy towards a desired sequence of mode transitions. The resulting Preferenced OGMP solves the loco-manipulation tasks effectively without any robot-specific reward shaping.   

\end{enumerate}

The rest of the paper is organized as follows:  Sec \ref{sec:ogmp_rev} revisits the OGMP theory, Sec \ref{sec:moha} introduces the hybrid automata perspective for oracles, Sec \ref{sec:method} presents the design methodology applied for the tasks of interest followed by the results in Sec \ref{sec:results}.

\section{Oracle Guided Multi-mode Policies Revisited}
\label{sec:ogmp_rev}
Given an oracle $\Xi$, which serves as a closed-loop state reference generator for solving a task $\mathcal{T}$, \cite{krishna2024ogmp} presents an oracle guided policy optimization framework to synthesize multi-mode policies to optimize a task objective $J_\mathcal{T}$. Specifically for any task variant  $\psi_\mathcal{T} \in \Psi_\mathcal{T}$, $\Xi$ provides a finite-horizon reference which is at most $\epsilon$ away from the optimal state trajectory $x^*_{t}$, from any given state $x_t$ based on feedback $\lambda_t$, as defined below.
\begin{subequations}
\begin{align}
 \quad & x^{\Xi}_{t:t+t_H} = \Xi(\lambda_t)\\
 \textrm{s.t.} 
 \quad &\exists \,\, x^{\Xi}_{t:t+t_H} 
 \quad \forall \, (x_t, \psi_{\mathcal{T}}) \\
 % \quad & x^{\Xi}_t = x\\
 \quad &\|x^{\Xi}_t - x^*_{t}\|_W<\epsilon \,\,\,\,\, \forall \, t\in [0,\,\infty)
\end{align}
\label{eq:oracle_def}
\end{subequations}
where $W$ is a user-defined weight matrix and $\epsilon$ is an unknown maximum deviation bound. With such a $\Xi$, OGMP localizes the exploration to a $\rho$-neighborhood of $x^\Xi$, ensuring effective learning for an appropriate choice of $\rho$. Formally,
\begin{subequations}
\vspace{-1mm}
\begin{align}
\quad & \pi^* := \arg \max_{\pi \in \Pi} J_\mathcal{T} \\
\textrm{s.t.}
\quad & \|x^{\pi}_t - x^{\Xi}_t\|_W< \rho \quad \forall t \in [0,\infty)
\end{align}
\label{eq:ogpo}
\end{subequations}
where $x^{\pi}$ are the states visited while rolling out policy $\pi$ and $\rho$ is the permissible state-bound. By bounded exploration, OGMP effectively escapes the multitude of local optima in the objective landscape, delivering the desired performance.  
\section{Multi-mode Oracles as Hybrid Automata}
\label{sec:moha}

This section establishes a functional view of multi-mode oracles as hybrid automata \cite{henzinger1996the}. Since oracles are reference generators, they have continuous dynamics that evolve the reference states and discrete jumps between the control modes based on environment feedback, leading to the following definition,  

\begin{equation}
        \Xi := (\mathcal{M}, \mathcal{X}, \Lambda , f , \mathcal{S})
\end{equation}

% \begin{aligned}
% \text{where} & \\
% & \mathcal{M} \text{ is the set of control modes} \\
% & \mathcal{X} \text{ is the continuous task/state space} \\
% & \Lambda \text{ is the continuous input space} \\
% & \quad \text{for environment feedback} \\
% & f: \mathcal{M} \times \mathcal{X} \times \Lambda \rightarrow \mathcal{X} \text{ is the reference generating} \\
% & \quad \text{dynamics}  \\
% & \mathcal{S} \text{ is a set of permissible transitions} \\
% & \quad \text{with reset, guard, and invariant conditions} \\
% \end{aligned}
where $\mathcal{M}$ is a discrete set of control modes, $\mathcal{X}$ is the continuous state space of the reference, $\Lambda$ is the continuous input space for environment feedback, $f: \mathcal{M} \times \mathcal{X} \times \Lambda \rightarrow \mathcal{X}$ is the reference generating dynamics, and $\mathcal{S}$ is a set of permissible transitions with reset, guard, and invariant conditions. Thus, $\Xi$ generates continuous references in $\mathcal{X}$ and discretely jumps between modes in $\mathcal{M}$ through switches in $\mathcal{S}$. 
% To design an oracle, the user can choose relevant $\mathcal{M}$ and $\mathcal{X}$ and design  $f$ and $\mathcal{S}$.
As an oracle, these design choices can be coarse, merely guiding optimization as an ansatz rather than offering a high-fidelity solution for the task.  
% Equivalently, $\Xi$ can also be perceived as finite directed multigraph vertices and edges ($\mathcal{M}$,$\mathcal{S}$), which helps visualization for design.  
We present the following example to elucidate this perspective. 

\textbf{Example - Reach-avoid Task:} We take the classical reach-avoid problem \cite{fisac2015reach} to demonstrate the design of an oracle for guided policy optimization. Consider a 2D robot modeled as a double integrator, $\ddot{q}_\text{robot} = u$, a parameterized controller, $ u = \pi(\,.\,;\theta)$ and a point obstacle and goal with generalized positions $q_\text{obstacles, }q_\text{goal} \in \mathbb{R}^2 \text{ respectively}$.
\begin{figure}[b]
\vspace{-2ex}
	\centering
	\includegraphics[width=0.4\textwidth]{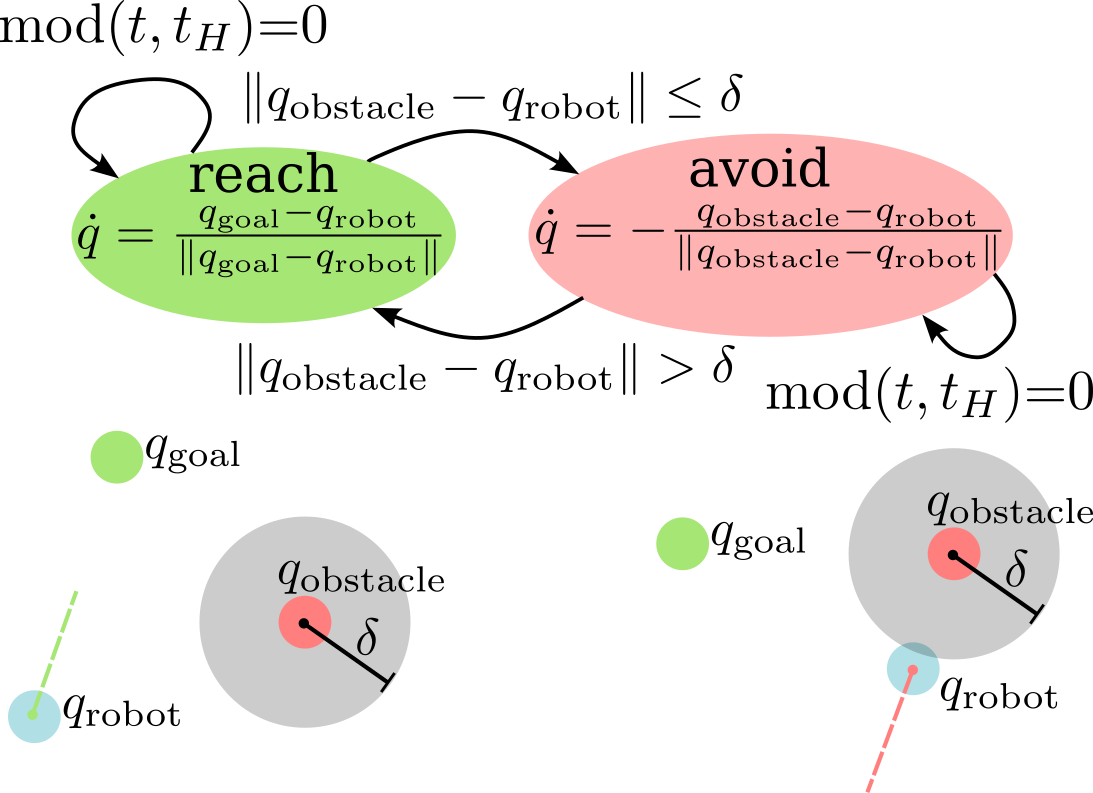}
	\caption{A multi-mode oracle designed as a hybrid automaton for the reach-avoid task. Dotted lines show the generated references in reach (green, left) and avoid (red, right) modes.}   
 % e) The closed-loop inference pipeline with the high-level oracle and the low-level multimodal policy 
	\label{fig:simple_example}
	% \vspace{-10mm}
\end{figure}
Given the current environment feedback $\lambda_t :=[q_\text{robot}, q_\text{obstacle},q_\text{goal},t]$, $\Xi$ generates a reference by integrating the continuous dynamics forward in time i.e., \[q^\Xi_\tau = \int_t^\tau \dot{q} dt, \quad \forall \tau \in [t,\,t+t_H] \] from the current mode in $\mathcal{M}:= \{\text{reach},\  \text{avoid}\} $. The switching conditions $\mathcal{S}$ ensure that when the obstacle is at least $\delta$ $m$ away, $\Xi$  remains in reach mode, generating references toward the goal, Fig. \ref{fig:simple_example} (left). Otherwise, it switches to avoid mode guiding the robot away from the obstacle, Fig. \ref{fig:simple_example} (right). A policy optimization algorithm is expected to learn the optimal $\theta$ by exploring the $\rho$-neighborhood of oracle's references as defined in (\ref{eq:ogpo}). We now highlight key remarks of this formulation. 

% \begin{remark}
% \label{remark:dyna_copling}
% $\lambda_t$ couples the environment and oracle's dynamics. Thus, depending on the control given by $\pi$ and the state of the environment, the sequence of mode transitions within the oracle can be different for different instances. Thus, while an oracle provides the set of all admissible mode switches, the policy assigns the mode-transition probabilities for a given environment.  
% \end{remark}
% \begin{remark}
% \label{remark:preference}
%     As a direct implication of Remark \ref{remark:dyna_copling}, one can now see a policy can converge to some undesirable mode-transition probabilities by exploring the dynamics. For instance, in the above example, $\pi$ can simply learn always to avoid an obstacle (stuck in the avoid $\rightarrow$ avoid ) transition infinitely) as opposed to going back to the reach mode and thereby reaching the desired goal. This calls for an explicit mechanism to enforce a desired preference of mode-transitions that solves a given task as later discussed in Sec \ref{subsec:preference}.  
% \end{remark}
% Here’s the revised text without the code box:

\begin{remark}
\label{remark:dyna_copling}
% $\lambda_t$ couples the environment dynamics with the oracle dynamics. Thus, depending on the control provided by $\pi$, the sequence of mode transitions within the oracle can differ between different environment rollouts. While the oracle defines the set of all admissible mode switches, the policy determines the mode-transition probabilities for a given environment.
$\lambda_t$ couples the environment dynamics with the oracle dynamics. Thus, depending on the control provided by $\pi$, the sequence of mode transitions within the oracle can vary across rollouts. While the oracle defines all the permissible mode switches, the policy determines the mode-transition probabilities.
\end{remark}

\begin{remark}
\label{remark:preference}
% As a direct implication of Remark \ref{remark:dyna_copling}, one can now observe that though $\Xi$ defines the admissible mode switches, the policy decides the transition probabilities through its actions for a given environment. Thus, an agent can converge to undesirable mode-transition probabilities by exploiting the dynamics and task objective. For instance, in the previous example, $\pi$ might learn to perpetually avoid an obstacle (getting stuck in an avoid $\rightarrow$ reach $\rightarrow$ avoid loop) rather than returning to the reach mode and, ultimately, the goal. This suggests the need for an explicit mechanism to enforce a preferred sequence of mode transitions that solve the task, as discussed later in Sec \ref{subsec:preference}.
From Remark \ref{remark:dyna_copling}, we note that an agent might converge to undesirable transition sequences by exploiting dynamics and task objectives. For example, $\pi$ may learn to avoid an obstacle indefinitely (getting stuck in an ``avoid $\rightarrow$ reach $\rightarrow$ avoid" loop) rather than reaching the goal and completing the task. This highlights the need for an explicit mechanism to enforce a preferred sequence of mode transitions, as discussed in Sec \ref{sec:preference}
\end{remark}

Thus, a hybrid automaton view of oracles provides a formal framework to design oracles systematically. Note that the notational complexity does not imply design complexity while scaling to high-dimensional systems, as we will demonstrate in Sec. \ref{sec:method}.  

\section{Methodology}
\label{sec:method}

\begin{figure}[t]
    \centering
    \includegraphics[width=0.48\textwidth]{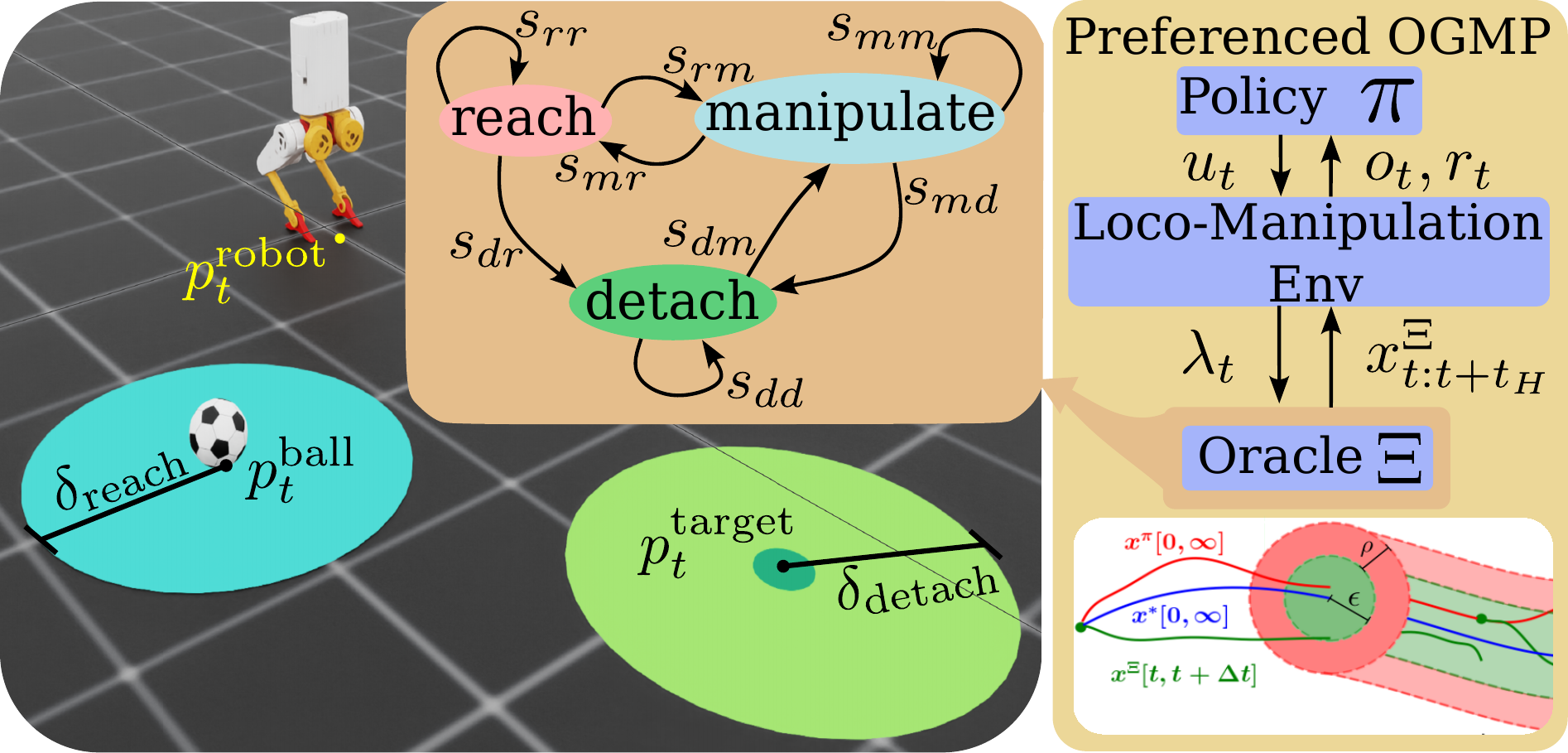}    
    \caption{Overview of the proposed framework: the environment dynamically queries the multi-mode oracle for a reference. A bounded exploration around the reference is then performed to learn dynamic loco-manipulation effectively.  
    }
    \label{fig:overview}
    \vspace{-5mm}
\end{figure}

This section presents our design methodology for solving uni-object bipedal loco-manipulation tasks. We first define the tasks of interest and the chosen modes of the oracle. Next, we outline the oracle design and propose a preferenced oracle-guided policy optimization, as shown in Fig. \ref{fig:overview}.

\subsection{Uni-object Loco-manipulation}
\label{subsec:uolm}
We divide the uni-object loco-manipulation task into three modes - reach, manipulate, and detach. With this $3$-mode abstraction, we aim to learn policies for multiple robots with varying dynamic and kinematic properties.  
We consider two tasks with different requirements in the finesse of manipulation: \emph{soccer} and \emph{move box}. The task variants are listed in Table \ref{tab:tasks}, along with the robots trained for each task.

\begin{table}[h]
    \renewcommand{\arraystretch}{1.3}
    \centering
 \caption{Variants of uni-object loco-manipulation tasks}   
    \begin{tabular}[0.48\textwidth]{|c|c|c|}
        \hline
        Tasks & Regularize & Robots \\
        \hline
        soccer-stop &$\times$  & HECTOR v1 \\
        soccer-kick & \checkmark  & Berkeley Humanoid, Unitree G1 and H1 \\
        move box & $\times$ & HECTOR v1 \\
        move box & \checkmark & Berkeley Humanoid, Unitree G1 and H1 \\
        \hline
    \end{tabular}
    \label{tab:tasks}
\end{table}

\subsubsection{Soccer task}
The robot is tasked with approaching a ball, dribbling it to a designated target, and detaching from it. The ball is arbitrarily initialized about a fixed distance from the robot, as seen in Fig. \ref{fig:overview} (left).
% The modes are reflected in the multi-mode oracle's hybrid automaton as the robot approaches the ball in reach mode and then transitions to manipulate mode, controlling and dribbling the ball toward the goal while maintaining balance.
Upon reaching the target position, the robot is expected to halt with two variations: \emph{stop} - bringing the ball to rest from omnidirectional dribbling, and \emph{kick} - shooting it towards the goal after unidirectional dribbling.
Guided by the oracle's coarse reference, the policy is expected to emergently learn contact-rich interactive behaviors to control the soccer ball.

\subsubsection{Move box task}
The move box task is analogous to soccer but only comprises the \emph{stop} variant for the detach mode. We choose boxes of dimensions comparable to each robot's nominal height to ensure whole-body loco-manipulation behaviors, continuously making and breaking contact with the box to ``negotiate" it towards the goal. 

\subsection{Oracle design for loco-manipulation}
\label{subsec:oracle_design}
We design a single oracle for all loco-manipulation tasks to provide references for the robot base and object states across different modes. Following (\ref{eq:oracle_def}), we define an oracle $\Xi$ with modes $\mathcal{M}:=\{$reach, manipulate, detach$\}$ and permissible mode transitions $\mathcal{S}$ as shown in Fig. \ref{fig:overview}. The oracle receives an environment feedback, $\lambda_t := [p_t^\text{robot}, p_t^\text{object}, p_t^\text{target}, t]$, where $p_t^\square$ denotes the position of $\square$, and $t$ is the current time in the episode. We chose the oracle reference horizon $t_H = 1$ sec and found the choice of $t_H$ does not affect performance, as shown in \cite{krishna2024ogmp}. 
% Before each policy step, the environment checks if any switching condition is active and queries $\Xi$ with $\lambda_t$ to generate a new reference. $\Xi$ determines the active mode and provides the corresponding $t_H$-length reference to the environment.
The reference state is defined as the CoM states of the robot and object, i.e.  $x^\Xi := [p^{\text{robot}}_{\Xi}, \theta^{\text{robot}}_{\Xi}, v^{\text{robot}}_{\Xi}, \omega^{\text{robot}}_{\Xi}, p^{\text{object}}_{\Xi}, v^{\text{object}}_{\Xi}]$, where $\theta$, $v$, and $\omega$ denote the orientation, linear velocity, and angular velocity respectively. 
All the references are generated as linear interpolations between the current and target positions. 
% Sophisticated references from model-based trajectory optimization can enhance policy performance \cite{krishna2024ogmp}, but generating such closed-loop references for dynamic tasks like dribbling is challenging.
Similar to warm-starting trajectory optimization \cite{winkler2018gait}, we found that linearly interpolated references were sufficient to solve our tasks as RL optimizes the policy subject to the full robot dynamics in the simulation. This property makes our approach simple and scalable for loco-manipulation to multiple robots, eliminating the need for dynamically feasible demonstrations \cite{sleiman2025guided} or carefully re-targeted kinematic references \cite{peng2020learning, li2024reinforcement}. 
By only having references for a subspace of the full state, the joint states are free for the policy to explore, thereby generating distinct behaviors for each task. 
% Note that the object references are only tracked in the detach modes (Table \ref{tab:rewards}), leaving room for emergent manipulation behaviors.
\begin{table}[b!]
\vspace{-1ex}
\renewcommand{\arraystretch}{1.8}
\caption{Rewards}
\label{tab:rewards}
\begin{tabular}{|m{16ex}m{2ex}c|}
    \hline
    \textbf{Term} & \textbf{Weight} & \textbf{Expression} \\
    \hline
    \multicolumn{3}{|l|}{\textbf{Task Rewards}} \\
    \hline
    Base Pos. &
    $0.3$ &
    $\exp(-5.0 \|p^{\text{robot}} - p^{\text{robot}}_{\Xi}\|)$ \\
    Base Ori. &
    $0.3$ &
    $\exp(-5.0 [1 - (\theta^{\text{robot}} \cdot \theta^{\text{robot}}_{\Xi})^2])$ \\
    Base Lin. Vel. &
    $0.15$ &
    $\exp(-2.0 \|v^{\text{robot}} - v^{\text{robot}}_{\Xi}\|)$ \\
    Base Ang. Vel. &
    $0.15$ &
    $\exp(-2.0 \|\omega^{\text{robot}} - \omega^{\text{robot}}_{\Xi}\|)$ \\
    Mode Preference &
    $-5.0$ &
    $\mathds{1} (m_t < \underset{0\leq \tau \leq t}{\max} ( m_\tau))$ \\
    Object Proximity &
    $0.5$ &
    $\mathds{1} (\|p^{\text{object}}- p^{\text{robot}}\| \leq h^\text{robot})\mathds{1}(m_t=1)$ \\
    Object Pos. &
    $1.0$ &
    $\exp(-2.0 \|p^{\text{object}} - p^{\text{object}}_{\Xi}\|) \mathds{1} (m_t=2)$ \\
    Object Lin. Vel. &
    $1.0$ &
    $\exp(-2.0 \|v^{\text{object}} - v^{\text{object}}_{\Xi}\|) \mathds{1} (m_t=2)$ \\
    Ball Rest Penalty &
    $-0.5$ &
    $\mathds{1} (\|v^{\text{object}}\| \leq 0.05 )\mathds{1}(m_t=1)$ \\
    \hline
    \multicolumn{3}{|l|}{\textbf{Regularization Rewards}} \\
    \hline
    Torque Mag. &
    $0.1$ &
    $\exp(-10.0 \|\tau\|)$ \\
    Torque Rate &
    $0.1$ &
    $\exp(-10.0 \|\tau - \tau_{\text{prev}}\|)$ \\
    Joint Vel. &
    $0.1$ &
    $\exp(-10.0 \|\dot{q}\|)$ \\
    Default Joint Pos. &
    $-0.1$ &
    $\sum_{i} |q_i - q_i^{\text{default}}|$ \\
    \hline 
\end{tabular}
\vspace{1ex}
\subcaption*{$\tau$, $q$, $\dot{q}$  denotes joint torques, positions and velocities, $h^\text{robot}$ denotes the nominal height of the robot and $\| \,.\, \|$ denotes the $l^2$-norm divided by corresponding normalizing constant.}
\end{table}
\subsection{Preferenced Oracle Guided Policy Optimization}
\label{sec:preference}
Our proposed preferenced oracle guided policy optimization framework for dynamic loco-manipulation is visualized in Fig. \ref{fig:overview}. At each environment step, the multi-mode oracle provides a $t_H$-long reference in the active mode determined by the environment feedback. The policy optimization is guided by the oracle's reference using permissible state terminations as in Section \ref{sec:ogmp_rev} and rewards in Table \ref{tab:rewards}. With this OGMP setup, we observed that the policy could exploit the automaton and converge to undesirable mode transition sequences. To overcome such local optima, we propose assigning mode preference as discussed below. 

\textbf{Mode preference
}:
Given a set of oracle modes and rewards, multiple solutions exist with some mode transitions resulting in undesirable behavior. For example, in Fig. \ref{fig:overview}, the agent can exploit the loop $s_{rr} \rightarrow s_{rm} \rightarrow s_{mr} \rightarrow s_{rr}$, leading to kicking the ball once at high velocity in the manipulate mode and continuously chasing it in the reach mode. Such ``creative" behaviors hinder finding the preferred behavior, such as $s_{rr} \rightarrow s_{rm} \rightarrow s_{mm} \rightarrow s_{md} \rightarrow s_{dd}$. Removing transitions (e.g., $s_{mr}$) to break the loop would be conservative as adversarial perturbations may drive the ball away from the robot, necessitating a switch to reach mode. Hence, we need to retain all necessary transitions but minimize the number of undesired transitions. Since it is infeasible to specify all optimal mode sequences apriori (as it is state-dependent), we propose assigning ranks to each mode. Ensuring increasing ranks point toward the preferred direction of transitions, we introduce the following preference reward : 
\begin{align*}
    r_{\text{preference}} = 
    \begin{cases} 
      -1 & m_t < \underset{0\leq \tau \leq t}{\max} ( m_\tau) \\
      0 & \text{otherwise}
    \end{cases}
\end{align*}

where $t$ is the current time step in the episode, and $m_t$ is the rank of the mode at $t$. Indicating preference as a penalizing reward allows yet minimizes transitions in the ``wrong direction" and results in subsequent recovery to the ``right direction". Specifically, we set ranks: reach-$0$, manipulate-$1$, and detach-$2$ and penalize any transition that decreases the mode rank compared to the highest rank observed thus far in the current episode. In the example above, switching from manipulate to reach (rank change: $2 \rightarrow 1$) is allowed but penalized, forcing the policy to learn the necessary recovery maneuvers. Note that the preference reward offers flexibility, allowing different modes to share the same rank and enabling multiple optimal behaviors. Unlike densely shaped reward terms, the above-proposed preference reward is sparse by definition and is left untuned across tasks. 

\begin{figure*}[t!]
	\centering
	\reflectbox{\includegraphics[width=\textwidth]{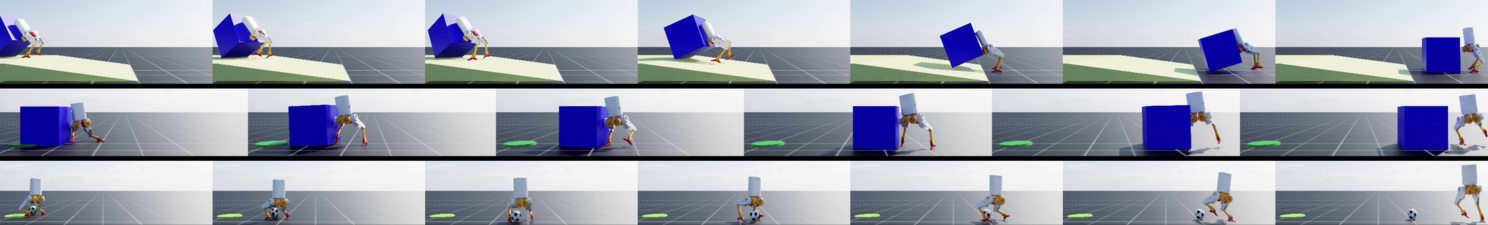}}
	\caption{Keyframes from the simulation: a) First row - moving the box along the slope. b) Second row - moving the box omnidirectionally along the plane c) Third row - dribbling a soccer ball in the soccer-stop task. In each case, the robot learns to manipulate the object to the target location (green circle).
 % through emergent behaviors using different parts of its morphology
 }  
	\label{fig:sim_results}
 \vspace{-5mm}
\end{figure*}

\textbf{Training details}: For training, we use the PPO \cite{schulman2017proximal} implementation of RSL RL \cite{rsl_rl} with the actor/critic network architectures being 1 RNN cell with 128 hidden units followed by an MLP with hidden units [200, 100]. The remaining hyperparameters are retained from the default configuration for RSL RL provided by Isaac Lab \cite{mittal2023orbit}. For all tasks specified in Table \ref{tab:tasks}, the policy observation comprises the proprioceptive robot states, relative distances to the object and target, and a phase indicating the current step in the reference. The policy outputs actuator PD targets, to be tracked by a low-level PD controller to compute torques. Our rewards are split into task and regularization terms, as seen in Table \ref{tab:rewards}. 
% Along with the base tracking from \cite{krishna2024ogmp}, object-tracking and simple task rewards are added. 
% The combination of dense tracking and sparse task rewards - such as object proximity and ball rest penalty, suffice to realize task-solving behaviors. 
Apart from the maximum episode length, we terminate based on the position errors of the robot and object relative to the current oracle reference.
For the permissible state bounds we use $\rho_x^{\text{robot}}=0.4$, $\rho_y^{\text{robot}}=0.4$ and $\rho_z^{\text{robot}}=0.2$ and $\rho_x^{\text{object}}=0.4$. To ensure realistic ball-robot dynamics in the soccer tasks, we incorporate a drag force represented by the equation $F_{drag} = - 0.5 v_{ball}^2$, similar to \cite{ji2023dribblebot}.

\section{Results}
\label{sec:results}

In this section, we present our results and performance analysis across the loco-manipulation tasks with different robots listed in Table \ref{tab:tasks}. We train on four robots that vary in morphology, actuation, and degrees of freedom - HECTOR v1, Berkeley Humanoid, Unitree G1, and H1. Since HECTOR is under active development, we train without regularization for simulation results. For other robots, we add regularization rewards considering the sim-to-real concerns, thus resulting in smoother motions. Our codebase with all the above robot and task variants is open-sourced \cite{ogmpisaacrepo}. Note that during inference, the policies are independent of the oracle, and the reference-based terminations are deactivated. 
% Thus, post-training, we use the oracle only as a tool for analysis. 
We present a qualitative overview followed by quantitative comparisons with relevant baselines. 
% We first analyze task performance, followed by the effect of mode preference reward, and compare our single multi-mode policy with the baseline as multiple uni-mode policies.

\subsection{Qualitative Analysis of Performance}

\begin{figure}
% \vspace{-5mm}
    \centering
    \includegraphics[width=0.48\textwidth]{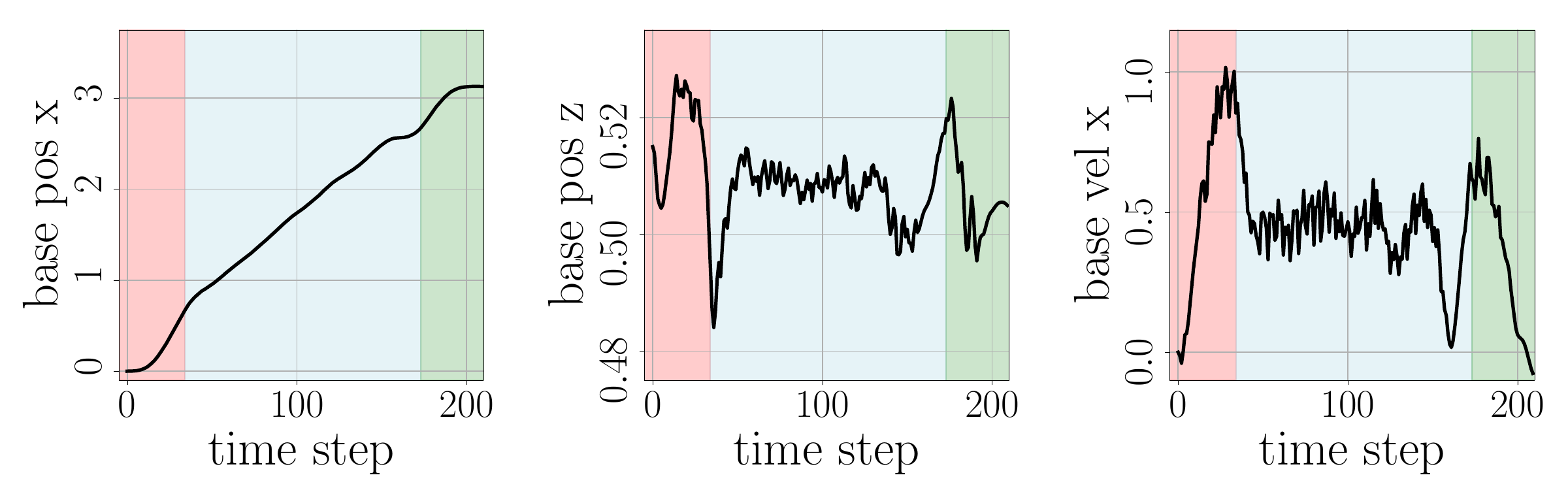}
    \caption{Selected CoM states in the soccer-kick task for Berkeley Humanoid. The colored regions indicate the active modes - reach (red), manipulate (blue), kick (green).}  
    \label{fig:base-pos-vel}
    \vspace{-5mm}
\end{figure}

\begin{figure*}[h]
    \centering
    \includegraphics[width=\textwidth]{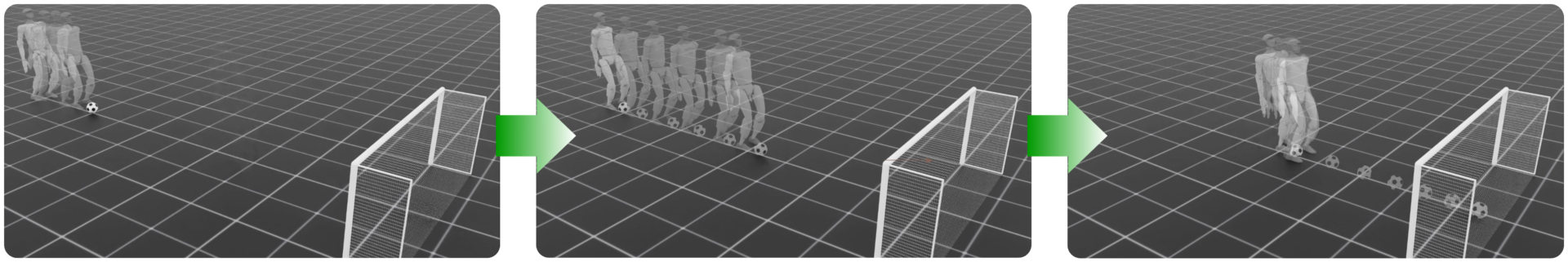}
    \caption{Motion trace of the policy trained with Unitree G1, transitioning through reach, manipulation, and detach modes.}
    \label{fig:g1-motion-trace}
    \vspace{-3mm}
\end{figure*}

As shown in Fig. \ref{fig:sim_results} and \ref{fig:g1-motion-trace}, the policies trained for each task successfully complete the intended loco-manipulation objective. 
Training a single multi-mode policy allows for implicit learning of non-trivial transition maneuvers. This is highlighted in Fig. \ref{fig:base-pos-vel} (left, center) for soccer-kick task, where the robot crouches and decelerates as it approaches the ball to prevent undesirable contacts. Similarly, before kicking, it accelerates to impart momentum to the ball, as shown in Fig. \ref{fig:base-pos-vel} (right).
% The mode-specific behaviors also demonstrate the advantage of allowing the policy to explore behaviors in the joint space while only providing crude references. 
As seen in the supplementary video, during the manipulate mode, HECTOR exhibits dribbling behaviors both when the ball is within and outside the support polygon spanned by its feet. It extensively uses both legs, shuffling the ball between its feet, kicking with one foot, and stopping it with the other. 
% Depending on the distance from the ball, the robot adjusts its height to exercise fine-grained control over the ball's movement. 
% In the stop mode, the robot crouches to trap and halt the ball using its calf and thigh. 
Despite using the same oracle and reward weights, we note the resulting policies for each robot differ in their strategies, adhering to the kinematic and dynamic limits. For instance, in soccer-kick, Berkeley Humanoid kicks using both legs, while G1 and H1 favor single-leg kicks. In the move box task, all robots employ their entire body to establish a ``stable" support plane on the object and manipulate it toward the target location. Rather than always facing the box, the policies prefer to exploit the robot morphology to cling to the object with non-trivial configurations. Hence, across all tasks, emergent behaviors involving contact-rich whole-body interactions are observed. This includes dynamic adjusting of contact points and non-trivial on-and-off contacts to stabilize and reactively ``negotiate" with the object. 
% the resulting policies differ in using each robot's unique morphologies. Since our approach framework separates the kinematic parameters from the <fill>, the transfer succeeds for both tasks. 
Thus, without additional robot-specific hyperparameter tuning, our approach enables successful policy synthesis across multiple robots and task variants within the abstraction of dynamic uni-object loco-manipulation.
% Next, we present quantitative analyses to justify our design choices.
% In the moving box over the slope task, the robot initially leaps to gain momentum, aiding in pushing the box uphill. Additionally, it tilts the box to reduce its contact with the plane until reaching the plateau.
% Empericlly , we noticed that the main parameter to choose was the speed of the robot. This relatively fixed the OGMP terminations and thresholds of the oracle. The relationship between the speed, terminations and thresholds carried over as the speed increased, showcasing the robustness of the method. Further quantitative analysis can reveal and strengthen this claim.

\subsection{Comparisons and Quantitative Analysis}
\subsubsection{Performance Metrics}

% We suspect the failures are due to the robot's morphology, where pushing in some directions is easier than others. For the soccer-omni task, we evaluate two metrics: the percentage of time the ball stays within a $0.5$m radius around the robot and the percentage of time the robot maintains contact with the ball in the manipulate mode. We observe that $95\%$ of the time, the robot successfully keeps the ball within $0.5$m of itself, and $68\%$ of the time, the robot maintains contact with the ball. This highlights the robot’s ability to engage in rich interaction dynamics without explicitly commanding contacts. This stands in contrast to model-based approaches, where maintaining ``stable" pre-scheduled contact $100\%$ of the time is required—a condition that is often unrealistic in environments with perturbations. By continuously making and breaking contact with the object, the robot is able to adjust its heading dynamically and ensure the successful completion of the task.
Table \ref{tab:metrics} presents the quantitative evaluation for the soccer-kick task. Over $100$ episodes, we randomize the initial joint states between $(-0.05, 0.05)$. We declare the episode to be successful when the ball reaches the goal and the robot stands within the detached region. The average ball contacts represent the number of contacts the robot makes with the ball in the manipulate mode. The fall percentage indicates the fraction of episodes in which the robot height falls below a threshold. Despite being trained without initial randomization, our approach succeeds in $98\%$ of the episodes with at least $9$ contacts made with the ball in the manipulate mode in each episode. We compute the resulting transition probabilities from each mode by counting the transitions and normalizing them over the total number of transitions. This provides the mode transition distributions the trained policy converged to, as shown in Fig. \ref{fig:trans_prob}. Self-transitions dominate the distribution due to the receding horizon design of oracles — having to stay in a mode until a different switching condition is met.

\begin{table}[h]
\renewcommand{\arraystretch}{1.3}    
\centering
\caption{Metrics over $100$ episodes for the soccer-kick task}
\begin{tabular}{|c|c|c|c|}
    \hline
    Policy & Success $\%$ & Avg. ball contacts & Fall $\%$ \\
    \hline
    multi-policy (baseline) & $26.0$ & $\textbf{12.46}$ & $69.0$ \\
    single-policy w/o pref & $28.0$ & $5.11$ & $\textbf{0.0}$ \\
    \textbf{single-policy (ours)} & $\textbf{98.0}$ & $9.76$ & $\textbf{0.0}$ \\
    \hline
\end{tabular}
\label{tab:metrics}
\vspace{-3mm}
\end{table}

\subsubsection{Effect of preference reward}

\begin{figure}[h]
    \centering
    \begin{subfigure}{0.23\textwidth}
        \centering
        \includegraphics[width=\textwidth]{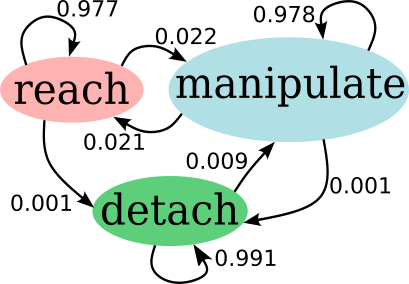}
        \caption{w/o preference reward}
        \label{fig:soc_wo_pref}
    \end{subfigure}    
    \hfill
    \begin{subfigure}{0.23\textwidth}
        \centering
        \includegraphics[width=\textwidth]{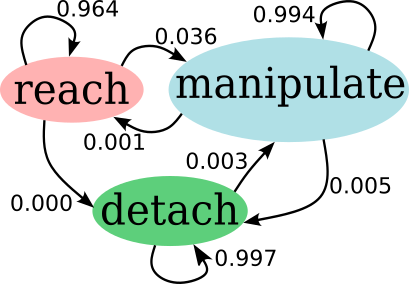}
        \caption{preference reward}
        \label{fig:soc_w_pref}
    \end{subfigure}
    
    \caption{Mode transition probabilities of the policy in the soccer-stop task.}
    \label{fig:trans_prob}
    \vspace{-3mm}
\end{figure}

We compare the effect of our proposed preference reward in increasing the number of desirable transitions as noted in Remark \ref{remark:preference}. From Fig. \ref{fig:trans_prob}.a, we observe that the policy without preference reward has comparable transition probabilities between reach $\rightarrow$ manipulate ($0.022$) and manipulate $\rightarrow$ reach ($0.021$). Thus, instead of an effective manipulation strategy, the policy exploits oracle dynamics, getting stuck in a reach $\rightarrow$ manipulate $\rightarrow$ reach loop, where the robot kicks the ball away and then chases it, as shown in the video. In contrast, the addition of the preference reward (Fig. \ref{fig:trans_prob}.b) ensures the transition probability from reach $\rightarrow$ manipulate is greater ($0.036$) than that of manipulate $\rightarrow$ reach ($0.001$). As shown in Table \ref{tab:metrics}, training with preference results in a higher number of ball contacts. While both policies have a low fall percentage, the preference-trained policy stands after kicking, while the non-preference policy avoids falling by perpetually walking behind the ball. Hence, adding preference results in $3$x more successful goal kicks than the one without, as seen in Table \ref{tab:metrics}. Thus, the preference reward helps disambiguate the desired optima corresponding to \emph{reach $\rightarrow$ manipulate $\rightarrow$ detach} from other locally optimal mode transitions.

\subsubsection{Single-policy vs. Multi-policy approaches}
\begin{figure}[t]
\centering
\includegraphics[width=0.48\textwidth]{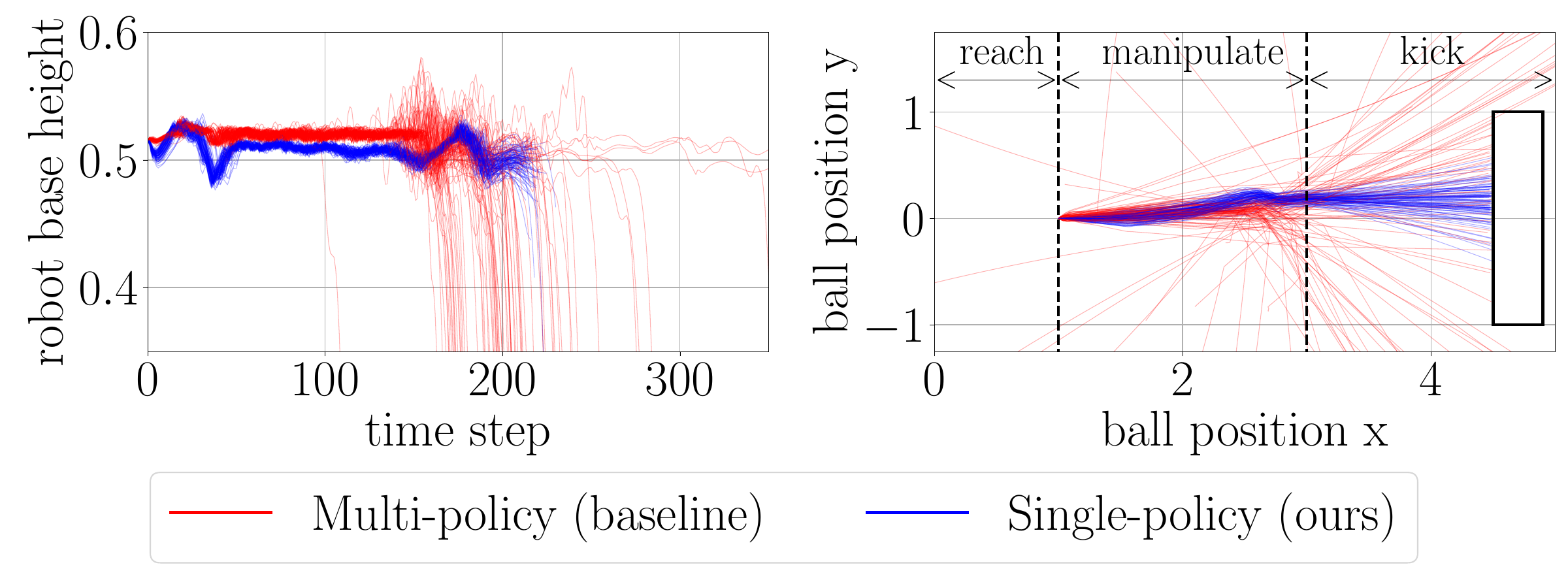}
\caption{Robot's base height and ball trajectories over $100$ episodes from randomized joint state initialization for the Berkeley Humanoid in the soccer-kick task}  
\label{fig:sp_mp_comparison}
\vspace{-3mm}
\end{figure}

We compare the performance of our single multi-mode policy with the established multiple uni-mode policies approach. As a baseline, we train three separate policies, one per mode in the soccer-kick task, using the same mode-specific rewards. During inference, we use the oracle as the finite state machine to switch between the policies similar to prior approaches \cite{dao2023simtoreallearninghumanoidbox, ji2023dribblebot}. When trained without initial randomization, the baseline policies fail to transition between each other, as seen in the attached video. To alleviate this, we randomize the initial floating base and joint states for the baseline manipulate and kick policies. As shown in Table \ref{tab:metrics}, the multi-policy baseline falls in $69\%$ of the episodes due to fragile inter-policy transitions, while our single-policy results in stabilizing base motions through robust inter-mode transitions, also seen in Fig. \ref{fig:sp_mp_comparison} (left). Comparing the dribbling performance, the baseline has $30\%$ more contacts during the manipulate mode. Despite the contact-rich manipulation, the multi-policy baseline fails to transition to the kick policy, as seen in  Fig. \ref{fig:sp_mp_comparison} (right). In contrast, having a comparable number of ball contacts, our approach results in a successful transition to kick mode and a task success rate of $98\%$ — $3$x more than the baseline approach. Despite having $0.3$x the learnable parameters of the multi-policy approach, our single multi-mode policy outperforms the baseline.

% Fig. \ref{fig:sp_mp_comparison} presents the qualitative behavior of the 
% two approaches using the robot and ball trajectories  

% Our single policy approach has a more 

% Likewise, the ball trajectory also follows the same trend. Despite being exclusively trained in kick mode, the multi-policy approach performs poorly, primarily due to undesired contact with the ball during policy transitions.

%  \begin{figure}[h]
% \centering
% \includegraphics[width=0.45\textwidth]{example-image-a}
% \caption{Soccer (kick) task for multiple robots - Berkeley Humanoid, Unitree G1 and Unitree H1}  
% \label{fig:multi-robot}
% \vspace{-3mm}
% \end{figure}

% \subsection{Ablation Studies}

% \begin{enumerate}
%     \item dribble bot-like rewards vs oracle guide
%     \item w/ and w/o preference rewards
%     \item given oracle, and speed .. show hyperparameter robustness on ( $\rho$ and  $\delta$, 
% \end{enumerate}

%  \begin{figure*}[t!]
% 	\centering
% 	\includegraphics[width=\textwidth]{}
% 	\caption{transitions probability oracles}  
% 	\label{fig:trans_prob}

% \end{figure*}

\section{Conclusion}
\label{sec:conclusion}

This work presents Preferenced Oracle Guided Multi-mode Policies and is applied to dynamic bipedal loco-manipulation. By designing reference-generating hybrid automata as coarse oracles with continuous reference dynamics and discrete mode switches, we effectively guide policy optimization. To minimize the number of undesired transition sequences, we introduce a novel mode preference reward, enhancing performance in tasks with multiple optimal mode transitions. The proposed approach was successfully validated over task variants like move box, soccer-stop, and soccer-kick with multiple robots of varying form factors, including HECTOR v1, Berkeley Humanoid, Unitree G1, and H1 in simulation. Compared with existing approaches, we demonstrate the advantage of training a single multi-mode policy with mode preference, resulting in non-trivial dynamic loco-manipulation. Ongoing work involves the sim-to-real transfer, with future efforts aimed at extending to more challenging tasks.

% I COMMENTED THIS TO MAKE THE REFERENCES FIT WITHOUT BREAKING

%%%%%%%%%%%%%%%%%%%%%%%%%%%%%%%%%%%%%%%%%%%%%%%%%%%%%%%%%%%%%%%%%%%%%%%%%%%%%%%%

\bibliographystyle{ieeetr}
\bibliography{references}

\begin{thebibliography}{10}

\bibitem{li2023multi}
J.~Li and Q.~Nguyen, ``Multi-contact mpc for dynamic loco-manipulation on humanoid robots,'' in {\em 2023 American Control Conference (ACC)}, pp.~1215--1220, 2023.

\bibitem{pickingup}
``Picking up momentum.''
\newblock \url{https://bostondynamics.com/blog/picking-up-momentum/}.

\bibitem{zhu2023efficient}
H.~Zhu, A.~Meduri, and L.~Righetti, ``Efficient object manipulation planning with monte carlo tree search,'' in {\em 2023 IEEE/RSJ International Conference on Intelligent Robots and Systems (IROS)}, pp.~10628--10635, 2023.

\bibitem{jorgensen202finding}
S.~J. Jorgensen, M.~Vedantam, R.~Gupta, H.~Cappel, and L.~Sentis, ``Finding locomanipulation plans quickly in the locomotion constrained manifold,'' in {\em 2020 IEEE International Conference on Robotics and Automation (ICRA)}, pp.~6611--6617, 2020.

\bibitem{sleiman2023versatile}
J.-P. Sleiman, F.~Farshidian, and M.~Hutter, ``Versatile multicontact planning and control for legged loco-manipulation,'' {\em Science Robotics}, vol.~8, no.~81, p.~eadg5014, 2023.

\bibitem{marcucci2024shortes}
T.~Marcucci, J.~Umenberger, P.~Parrilo, and R.~Tedrake, ``Shortest paths in graphs of convex sets,'' {\em SIAM Journal on Optimization}, vol.~34, p.~507–532, Feb. 2024.

\bibitem{lee2020learning}
J.~Lee, J.~Hwangbo, L.~Wellhausen, V.~Koltun, and M.~Hutter, ``Learning quadrupedal locomotion over challenging terrain,'' {\em Science robotics}, vol.~5, no.~47, p.~eabc5986, 2020.

\bibitem{cheng2024extreme}
X.~Cheng, K.~Shi, A.~Agarwal, and D.~Pathak, ``Extreme parkour with legged robots,'' in {\em 2024 IEEE International Conference on Robotics and Automation (ICRA)}, pp.~11443--11450, IEEE, 2024.

\bibitem{radosavovic2024real}
I.~Radosavovic, T.~Xiao, B.~Zhang, T.~Darrell, J.~Malik, and K.~Sreenath, ``Real-world humanoid locomotion with reinforcement learning,'' {\em Science Robotics}, vol.~9, no.~89, p.~eadi9579, 2024.

\bibitem{tan2018sim}
J.~Tan, T.~Zhang, E.~Coumans, A.~Iscen, Y.~Bai, D.~Hafner, S.~Bohez, and V.~Vanhoucke, ``Sim-to-real: Learning agile locomotion for quadruped robots,'' {\em arXiv preprint arXiv:1804.10332}, 2018.

\bibitem{hwangbo2019learning}
J.~Hwangbo, J.~Lee, A.~Dosovitskiy, D.~Bellicoso, V.~Tsounis, V.~Koltun, and M.~Hutter, ``Learning agile and dynamic motor skills for legged robots,'' {\em Science Robotics}, vol.~4, no.~26, p.~eaau5872, 2019.

\bibitem{margolis2024rapid}
G.~B. Margolis, G.~Yang, K.~Paigwar, T.~Chen, and P.~Agrawal, ``Rapid locomotion via reinforcement learning,'' {\em The International Journal of Robotics Research}, vol.~43, no.~4, pp.~572--587, 2024.

\bibitem{siekmann2021blind}
J.~Siekmann, K.~Green, J.~Warila, A.~Fern, and J.~Hurst, ``{Blind Bipedal Stair Traversal via Sim-to-Real Reinforcement Learning},'' in {\em RSS}, 2021.

\bibitem{miki2022learning}
T.~Miki, J.~Lee, J.~Hwangbo, L.~Wellhausen, V.~Koltun, and M.~Hutter, ``Learning robust perceptive locomotion for quadrupedal robots in the wild,'' {\em Science Robotics}, 2022.

\bibitem{chen2023visual}
T.~Chen, M.~Tippur, S.~Wu, V.~Kumar, E.~Adelson, and P.~Agrawal, ``Visual dexterity: In-hand reorientation of novel and complex object shapes,'' {\em Science Robotics}, vol.~8, no.~84, p.~eadc9244, 2023.

\bibitem{dadiotis2025dynamicobjectgoalpushing}
I.~Dadiotis, M.~Mittal, N.~Tsagarakis, and M.~Hutter, ``Dynamic object goal pushing with mobile manipulators through model-free constrained reinforcement learning,'' 2025.

\bibitem{jeon2024learning}
S.~Jeon, M.~Jung, S.~Choi, B.~Kim, and J.~Hwangbo, ``Learning whole-body manipulation for quadrupedal robot,'' {\em IEEE Robotics and Automation Letters}, vol.~9, no.~1, pp.~699--706, 2024.

\bibitem{huang2022creating}
X.~Huang, Z.~Li, Y.~Xiang, Y.~Ni, Y.~Chi, Y.~Li, L.~Yang, X.~B. Peng, and K.~Sreenath, ``Creating a dynamic quadrupedal robotic goalkeeper with reinforcement learning,'' 2022.

\bibitem{ji2023dribblebot}
Y.~Ji, G.~B. Margolis, and P.~Agrawal, ``Dribblebot: Dynamic legged manipulation in the wild,'' in {\em 2023 IEEE International Conference on Robotics and Automation (ICRA)}, pp.~5155--5162, 2023.

\bibitem{dao2023simtoreallearninghumanoidbox}
J.~Dao, H.~Duan, and A.~Fern, ``Sim-to-real learning for humanoid box loco-manipulation,'' in {\em 2024 IEEE International Conference on Robotics and Automation (ICRA)}, pp.~16930--16936, 2024.

\bibitem{haarnoja2024learning}
T.~Haarnoja, B.~Moran, G.~Lever, S.~H. Huang, D.~Tirumala, J.~Humplik, M.~Wulfmeier, S.~Tunyasuvunakool, N.~Y. Siegel, R.~Hafner, M.~Bloesch, K.~Hartikainen, A.~Byravan, L.~Hasenclever, Y.~Tassa, F.~Sadeghi, N.~Batchelor, F.~Casarini, S.~Saliceti, C.~Game, N.~Sreendra, K.~Patel, M.~Gwira, A.~Huber, N.~Hurley, F.~Nori, R.~Hadsell, and N.~Heess, ``Learning agile soccer skills for a bipedal robot with deep reinforcement learning,'' {\em Science Robotics}, vol.~9, no.~89, p.~eadi8022, 2024.

\bibitem{sleiman2025guided}
J.~P. Sleiman, M.~Mittal, and M.~Hutter, ``Guided reinforcement learning for robust multi-contact loco-manipulation,'' in {\em Proceedings of The 8th Conference on Robot Learning} (P.~Agrawal, O.~Kroemer, and W.~Burgard, eds.), vol.~270 of {\em Proceedings of Machine Learning Research}, pp.~531--546, PMLR, 06--09 Nov 2025.

\bibitem{krishna2024ogmp}
L.~Krishna, N.~Sobanbabu, and Q.~Nguyen, ``Ogmp: Oracle guided multi-mode policies for agile and versatile robot control,'' 2024.

\bibitem{henzinger1996the}
T.~Henzinger, ``The theory of hybrid automata,'' in {\em Proceedings 11th Annual IEEE Symposium on Logic in Computer Science}, pp.~278--292, 1996.

\bibitem{fisac2015reach}
J.~F. Fisac, M.~Chen, C.~J. Tomlin, and S.~S. Sastry, ``Reach-avoid problems with time-varying dynamics, targets and constraints,'' in {\em Proceedings of the 18th International Conference on Hybrid Systems: Computation and Control}, HSCC '15, (New York, NY, USA), p.~11–20, Association for Computing Machinery, 2015.

\bibitem{winkler2018gait}
A.~W. Winkler, C.~D. Bellicoso, M.~Hutter, and J.~Buchli, ``Gait and trajectory optimization for legged systems through phase-based end-effector parameterization,'' {\em IEEE Robotics and Automation Letters}, vol.~3, no.~3, pp.~1560--1567, 2018.

\bibitem{peng2020learning}
X.~B. Peng, E.~Coumans, T.~Zhang, T.-W. Lee, J.~Tan, and S.~Levine, ``{Learning Agile Robotic Locomotion Skills by Imitating Animals},'' in {\em Proceedings of Robotics: Science and Systems}, (Corvalis, Oregon, USA), July 2020.

\bibitem{li2024reinforcement}
Z.~Li, X.~B. Peng, P.~Abbeel, S.~Levine, G.~Berseth, and K.~Sreenath, ``Reinforcement learning for versatile, dynamic, and robust bipedal locomotion control,'' {\em The International Journal of Robotics Research}, vol.~0, no.~0, p.~02783649241285161, 0.

\bibitem{schulman2017proximal}
J.~Schulman, F.~Wolski, P.~Dhariwal, A.~Radford, and O.~Klimov, ``Proximal policy optimization algorithms,'' {\em arXiv preprint arXiv:1707.06347}, 2017.

\bibitem{rsl_rl}
``rsl\_rl: Fast and simple implementation of rl algorithms, designed to run fully on gpu.''
\newblock \url{https://github.com/leggedrobotics/rsl_rl}.

\bibitem{mittal2023orbit}
M.~Mittal, C.~Yu, Q.~Yu, J.~Liu, N.~Rudin, D.~Hoeller, J.~L. Yuan, R.~Singh, Y.~Guo, H.~Mazhar, A.~Mandlekar, B.~Babich, G.~State, M.~Hutter, and A.~Garg, ``Orbit: A unified simulation framework for interactive robot learning environments,'' {\em IEEE Robotics and Automation Letters}, vol.~8, no.~6, pp.~3740--3747, 2023.

\bibitem{ogmpisaacrepo}
\url{https://github.com/DRCL-USC/ogmp_isaac}.

\end{thebibliography}

\end{document}